\documentclass[runningheads]{llncs}
\usepackage{graphicx}

\usepackage{tikz}
\usepackage{comment}
\usepackage{amsmath,amssymb} 
\usepackage{color}
\usepackage{mathtools}
\usepackage{siunitx}
\usepackage{floatrow}
\usepackage{hyperref}
\usepackage{booktabs}
\usepackage{tabularx}
\usepackage{float}
\restylefloat{figure}

\usepackage[accsupp]{axessibility}  

\usepackage[a4paper, margin=1.3in]{geometry}

\newcommand{\mc}[1]{\mathcal{#1}}

\newcommand{\inp}{\mathbb{R}^{d\times d\times c_\text{in}}}
\newcommand{\tar}{\mathbb{R}^{d\times d\times c_\text{out}}}
\newcommand{\norm}[1]{\left\lVert#1\right\rVert}

\newcommand{\outdom}{\mathcal{Y}}
\newcommand{\wass}{\mathcal W}

\begin{document}

\pagestyle{headings}
\mainmatter
\def\ECCVSubNumber{0005}  

\title{Unsupervised Joint Image Transfer and Uncertainty Quantification Using Patch Invariant Networks } 

\titlerunning{Unsupervised Joint Image Transfer and Uncertainty Quantification}
%
\author{Christoph Angermann \and
Markus Haltmeier \and
Ahsan Raza Siyal}
\authorrunning{C. Angermann et al.}
%
\institute{Department of Mathematics, University of Innsbruck, Technikerstraße 13, 6020 Innsbruck, Austria\\
\url{applied-math.uibk.ac.at}}
\maketitle

\begin{abstract}
Unsupervised image transfer enables intra- and inter-modality image translation in applications where a large amount of paired training data is not abundant.
To ensure a structure-preserving mapping from the input to the target domain, existing methods for unpaired image transfer are commonly based on cycle-consistency, causing additional computational resources and instability due to the learning of an inverse mapping. This paper presents a novel method for uni-directional domain mapping that does not rely on any paired training data. A proper transfer is achieved by using a GAN architecture and a novel generator loss based on patch invariance. To be more specific, the generator outputs are evaluated and compared at different scales, also leading to an increased focus on high-frequency details as well as an implicit data augmentation. 
This novel patch loss also offers the possibility to accurately predict aleatoric uncertainty by modeling an input-dependent scale map for the patch residuals.
The proposed method is comprehensively evaluated on three well-established medical databases. 
As compared to four state-of-the-art methods, we observe significantly higher accuracy on these datasets, indicating great potential of the proposed method for unpaired image transfer with uncertainty taken into account.
Implementation of the proposed framework is released here: \url{https://github.com/anger-man/unsupervised-image-transfer-and-uq}.

\keywords{unsupervised image transfer, uncertainty quantification, generative adversarial network, patch invariance, modality propagation, accelerated MRI, radiotherapy}
\end{abstract}

\section{Introduction}

Image transfer, e.g., within and across medical imaging modalities, has gained a lot of popularity in the last decade of research \cite{wang2020}. The application range of inter- and intra-modality image translation is multifaceted and can help to overcome key weaknesses of an acquisition method. 
For example, modality propagation in magnetic resonance imaging (MRI) is of high interest since acquisition of multiple contrasts is crucial for better diagnosis for many clinical protocols \cite{upadhyay2021}. Especially acquiring T1-weighted (T1w) and T2-weighted (T2w) contrasts increases scanning time significantly and thus is often formulated as a translation task from T1w to T2w.
Another example is accelerated MRI, where medical costs and patient stress are minimized by decreasing the amount of k-space measurements \cite{fastmri,hyun2018}. Such methods for MRI reconstruction of undersampled measurements allow the use of MRI in applications where it is currently too time and resource intensive. A third application to mention here is automated computed tomography (CT) synthesis based on MRI images, which allows for MRI-only treatment planning in radiation therapy. A MRI-to-CT synthesizing method can eliminate the need for CT simulation and therefore improves the treatment workflow and reduces radiation exposure for the patient during radiotherapy \cite{wang2020,lei2019,yang2018,wolterink2017,hiasa2018}.

Just discussed applications correspond to (un)supervised image transfer, which targets at translating an image from one domain to another. Supervised approaches exploit the inter-domain correspondence between input and output data \cite{upadhyay2021robustness}. These methods rely on large amounts of paired data and perfectly registered scans of the same patient, which are not always abundant in medical applications \cite{lei2019}. 
Unsupervised approaches are commonly build on a generative adversarial network (GAN) \cite{goodfellow2014} that assimilates the distribution of the generated samples to the real distribution of the target domain by employing an adversarial discriminator network. To ensure that the synthesized output does not become irrelevant to the input, additional constraints may be added to the generator loss \cite{zhu2017,fu2019,benaim2017,park2020}.
 Especially cycle-consistency is a well-received method for structure preservation in fully unsupervised medical image transfer \cite{yang2018,wolterink2017,hiasa2018,upadhyay2021robustness}. However, a cycle-consistent GAN (cycleGAN) requires the parallel learning of an inverse mapping. As a result, the training time is significantly increased and the final performance depends on the inverse transfer function. Furthermore, cycle-consistent GANs compare the reconstruction on a whole-image base and therefore pay less attention to fine structures and high-frequency details.

Although current methods provide powerful tools for high-dimensional image transfer, the generator loss calculation usually assumes that the learned mappings are correct. This represents a significant source for instability and erroneous predictions when considering out-of-distribution (OOD) data during training \cite{upadhyay2021robustness,kendall2017}. Tackling data-dependent uncertainty in deep computer vision has raised a lot of interest in the last years of research and provided effective tools to check the reliability of a model's predictions in supervised applications. However, research on modeling uncertainty inherent from data in a completely unsupervised setting is still limited \cite{upadhyay2021robustness,saatci2017} and needs a deeper investigation.\\

This paper presents a novel GAN approach to fully unpaired medical image transfer, including prediction of data-dependent uncertainty and invariance over patches. To be more exact, a Wasserstein generative adversarial network (WGAN) \cite{arjovsky2017} is leveraged to an uni-directional image transfer model. Structural correspondence between input and target modality  is guaranteed using a novel generator loss that enforces invariance over image patches. Furthermore, the patch-based residuals are assumed to follow a zero-mean Laplace distribution with the scale parameter being a function of the input. As a consequence, the generator is allowed to predict uncertainty that operates as a learned loss attenuation and can be used to indicate the quality of a transferred image in the absence of ground truth data. The proposed model and training strategy is evaluated in three different unsupervised scenarios: modality propagation using T1w and T2w brain MRI from the IXI \cite{ixi} database; accelerated MRI enhancement using emulated single-coil knee MRI from the FastMRI \cite{fastmri} database; MRI-to-CT synthesis using head CT scans from the CQ500 \cite{cq500} database .
The proposed framework is benchmarked against state-of-the-art works for uni-directional and bi-directional image translation \cite{upadhyay2021robustness,zhu2017,fu2019,park2020}. We not only evaluate accuracy on unseen test data but further investigate robustness to perturbed inputs. \\

\textbf{Contributions}:
\begin{itemize}
	\item We present an unidirectional framework that enables fully unsupervised image transmission of medical data while preserving fine structures.
	\item Structural correspondence between different characteristics and modalities is ensured by an improved generator loss based on patch invariance. This also yields implicit data augmentation for the critic and generator networks.
	\item In addition to the image transmission, the model provides an uncertainty map that correlates with the prediction error, indicating the quality of a mapped instance.
\end{itemize}

\section{Related Work}
\subsection{Generative Adversarial Networks}
The GAN architecture \cite{goodfellow2014} is composed of a generator network $G:\mc Z \to \mc X$ and an adversarial part $f:\mc X\to[0,1]$. The generator maps from a latent space $\mc Z$ to image space $\mc X$, where the parameters of $G$ are adapted such that the distribution of the synthesized  examples assimilates to the real data distribution on $\mc X$. Simultaneously, the adversarial $f$ is trained to distinguish between synthesized and real instances. In a two-player min-max game, generator parameters are updated to fool a steadily improving discriminator \cite{angermann2022}. Improving the joint loss functional of the generating and the adversarial part yielded improved modifications of the initial GAN framework, like WGAN \cite{arjovsky2017}, improved WGAN \cite{gulrajani2017}, LSGAN \cite{mao2017} or SNGAN \cite{miyato2018}. While GANs reach outstanding performance in image synthesis \cite{karras2019,karras2018}, they are also well accepted in improving prediction quality in supervised image applications such as super-resolution \cite{ledig2017}, paired image translation \cite{isola2017,arslan2019}, and medical image enhancement \cite{lei2019,armanious2020}.

\subsection{Unpaired Image Transfer \& Domain Mapping}
\label{sec:sota}
Unpaired image translation maps an image from input to target domain where corresponding samples from both spaces are hard to obtain or applied registration methods yield too much misalignment.  In these cases, cycleGAN \cite{zhu2017} has become the gold standard since it learns an inverse mapping from target domain back to input space. The core idea of cycleGAN is that the synthesized image must retain enough detail of the input instance in the target domain to allow for reconstruction. Especially in the medical sector, a structure-preserving transfer function is of high priority. Wolterink et al. \cite{wolterink2017} utilized cycle-consistency for MRI-only  treatment planning in radiotherapy. Hiasa et al. \cite{hiasa2018} improved the cycleGAN architecture in this application area by adding a gradient consistency loss to pay more attention to the edges in the image. Yang et al. \cite{yang2018} added to cycleGAN a structure-consistency loss based on a modality independent neighborhood descriptor.

Learning an inverse GAN framework simultaneously (bi-directional) in order to ensure input-output consistency increases hardware requirements and introduces an additional instability if the inverse generator is not trained sufficiently or if the transfer mapping is not injective. Fu et al. \cite{fu2019} investigated geometry-consistent GAN (gcGAN), an uni-directional approach that enforces consistency when applying geometric transformations (rotation, flipping) before and after propagation through the generator network. Benaim et Wolf \cite{benaim2017} considered GAN in combination with a distance constraint, where the distance between two samples from the input domain should be preserved after mapping to the target domain. A very recent and successful  uni-directional approach to unpaired domain mapping was made by Park et al. \cite{park2020} that uses contrastive unpaired translation (CUT), i.e., structure-consistency is preserved by matching patches of the input and the synthesized instance using an additional classification step.
\subsection{Uncertainty Quantification}
\label{sec:uc}
Uncertainty quantification methods have been applied to solve a variety of real-world problems in computer vision where in addition to the model's response also a measure on its confidence is provided \cite{abdar2021}. In general, two broad categories of uncertainty are considered: aleatoric uncertainty captures noise inherent in the data and epistemic/model uncertainty describes uncertainty in the model parameters \cite{kendall2017}. The latter type of uncertainty occurs in finite data settings and thus can be explained away providing a sufficient amount of data. Bayesian models provide a mathematically grounded framework that can account for model uncertainty in combination with Bayesian inference techniques.
Gal et Ghahramani \cite{gal2016} set up a theoretical framework that casts the dropout technique as approximate Bayesian inference, enabling a rather simple calculation of epistemic uncertainty by multiple network forward passes. 
The works of Saatci et Wilson \cite{saatci2017} as well as  Palakkadavath et Srijith \cite{palakkadavath2021} leverage this framework to Bayesian GANs and show that considering Bayesian learning principles can address mode collapse in image synthesis.
 Kendall et Gal \cite{kendall2017} have explored the benefits of modeling aleatoric and epistemic uncertainty simultaneously in image segmentation and regression and concluded that the two types of uncertainty are not mutually exclusive, but in fact complementary in different data scenarios. Upadhyay et al. modeled aleatoric uncertainty for MRI image enhancement \cite{upadhyay2021} and unsupervised image transfer \cite{upadhyay2021robustness} by introducing uncertainty-aware generalized adaptive cycleGAN (UGAC). Therefore, the latter work will also serve as a benchmark method for the proposed uncertainty-aware uni-directional image transfer approach.

\section{Method}
\label{sec:method}

\subsection{Preliminaries \& GAN Architecture}

The underlying structure of the proposed uncertainty-aware domain mapping is a GAN combined with a patch invariant generator term. Let $\mc X\subset \inp$ and $\mc Y\subset \tar$ denote the input and the target domain, respectively. For simplicity we consider quadratic instances with the number of image pixels equal to $d^2$. Furthermore, let $X\coloneqq \{x_1,\ldots,x_M\}$ be the set of $M$ given input images and $Y\coloneqq\{y_1,\ldots,y_N\}$ the set of $N$ available but unaligned target images.
$P_\mc{X}$ and $P_\mc{Y}$ denote the distributions of the images in both domains.
The proposed image transfer is built on a generator function $G_{\theta}\colon\mc X\to \mc Y$, which aims to map an input sample to a corresponding instance in the target domain. The generator function is approximated by a convolutional neural network (CNN), which is parameterized by a weight vector $\theta$.
By adjusting $\theta$, the distribution $P_{\theta}$ of generator outputs may be brought closer to the real data distribution $P_\mc{Y}$ in the target domain.
  The distance between the generator distribution and the real distribution is estimated with the help of the critic $f_\omega\colon\mc Y \to \mathbb{R}$, which is parameterized by weight vector $\omega$ and is trained simultaneously with the generator network since $P_\theta$ changes after each update to the generator weights $\theta$ \cite{angermann2022}. 

%
%
We choose a network critic based on the Wasserstein-1 distance \cite{arjovsky2017,angermann2022,villani2008}.
The Wasserstein-1 distance between two distributions $P_1$ and $P_2$ is defined as
$ \wass_1(P_1, P_2) \coloneqq \inf_{J\in\mathcal J(P_1,P_2)}\mathbb{E}_{(x,y)\sim J}\norm{x-y}$,
where the infimum is taken over the set of all joint probability distributions that have marginal distributions $P_1$ and $P_2$. 
The Kantorovich-Rubinstein duality \cite{villani2008} yields
\begin{align}\label{eq:wgan}
\wass_1(P_1,P_2)
=\sup_{\norm{f}_L\leq 1}\left[\underset{y\sim P_1}{\mathbb E}f(y)- \underset{y\sim P_{2}}{\mathbb{E}}f(y)\right],
\end{align}
where $\norm{\cdot}_L\leq C$ denotes that a function is $C$-Lipschitz. Equation (\ref{eq:wgan}) indicates that a good approximation to $\wass_1(P_\outdom,P_\theta)$ is found by maximizing ${\mathbb E}_{y\sim P_\outdom}f_\omega(y)- {\mathbb{E}}_{y\sim P_\theta}f_\omega(y)$ over the set of CNN weights $\{\omega\mid f_\omega\colon\outdom\to\mathbb{R}\ \text{1-Lipschitz}\}$, where the Lipschitz continuity of $f_\omega$ can be enhanced via a gradient penalty \cite{gulrajani2017}. Given training batches $\mathbf{y}=\{y_n\}_{n=1}^b,\ y_n \overset{\mathrm{iid}}{\sim} P_\mc{Y}$ and $\mathbf{x}=\{x_n\}_{n=1}^b,\ x_n\overset{\mathrm{iid}}{\sim} P_\mc{X}$, this yields the following  empirical risk for critic $f_\omega$:
\begin{equation}
\footnotesize
\label{eq:critic}
\begin{split}
\ell_\text{cri}(\omega,\theta,\mathbf{y},\mathbf{x},p)\coloneqq \frac{1}{b}\sum_{n=1}^{b}f_\omega(G_\theta(x_n))-f_\omega(y_n)
+p\cdot \left(\left(\norm{\nabla_{\tilde y_n}f_\omega(\tilde y_n)}_2-1\right)_+ \right)^2,
\end{split}
\end{equation}
where $p$ denotes the influence of the gradient penalty, $( \cdot) _+\coloneqq \max(\{0,\cdot\})$  and $\tilde y_n \coloneqq \epsilon_n\cdot G_\theta(x_n)+ (1-\epsilon_n)\cdot y_n$ for $\epsilon_n\overset{\mathrm{iid}}{\sim} \mc U[0,1]$. Since only the first term of the functional in (\ref{eq:critic}) depends on $\theta$ and the goal for the generator is to minimize the Wasserstein-1 distance, the adversarial empirical risk for generator $G_\theta$ simplifies as follows:
\begin{align}
\label{eq:gen}
\ell_\text{gen}(\theta,\omega,\mathbf{x})\coloneqq -\frac{1}{b}\sum_{n=1}^{b}f_\omega(G_\theta(x_n)).
\end{align}

\subsection{Patch Invariance}
\begin{figure}[t!]
	\centering
	\includegraphics[width=.978\textwidth]{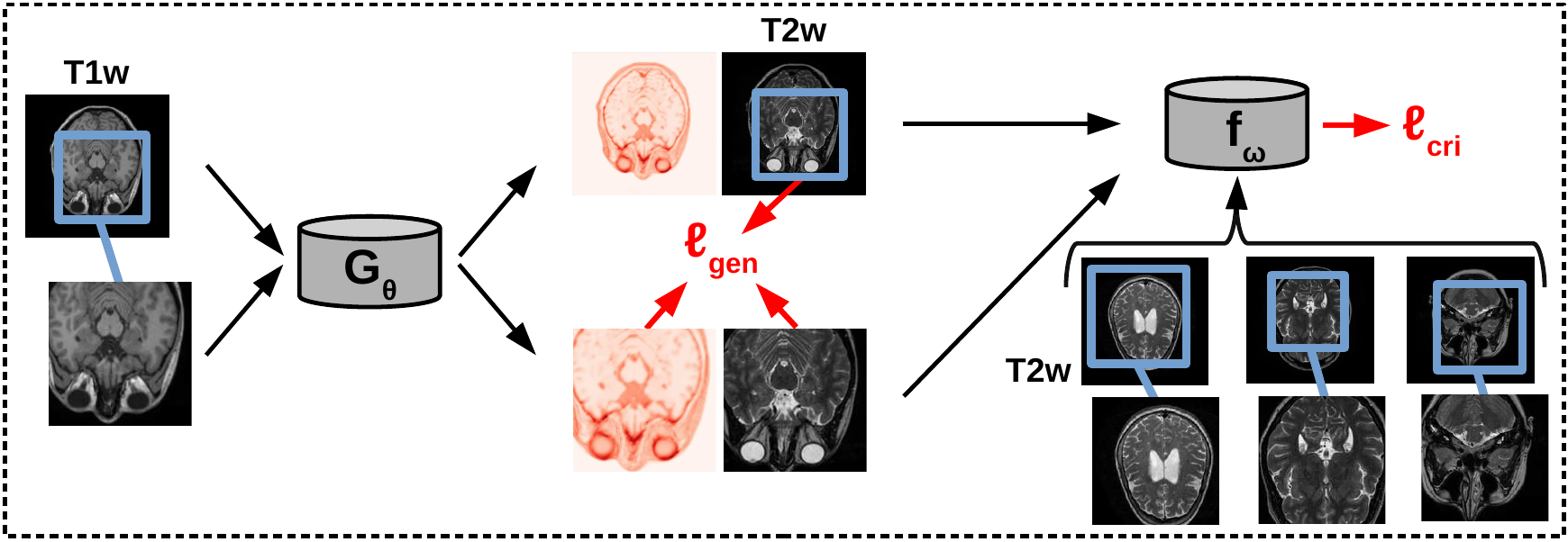}
	\caption{Utilizing patch invariance for unsupervised MRI propagation and uncertainty quantification. The T1w input and a corresponding random patch on a finer scale are fed to the generator $G_\theta$, which outputs the synthesized T2w counterparts  and corresponding scale maps (red). The synthesized patch and the corresponding patch of the full-size output are compared. Loss attenuation is introduced by the scale map of the synthesized patch. The generator is additionally updated using the Wasserstein-1 distance, estimated with the help of $f_\omega$.}
	\label{fig:model}
\end{figure}

In the frame of medical image translation, it is not sufficient to ensure that the output samples lie in the target domain. Great attention should be paid that a model also preserves global structure as well as fine local details.
Let $x\in\mathbb{R}^{d\times d\times c}$ and $\Phi \coloneqq \left\{(\rho,j_1,j_2)\in[0.7,1]\times [0,d]^2 \ \big\vert\  j_1+\rho d\leq d \wedge  j_2+\rho d \leq d\right\}$. We define the patch operator $\mc P:\Phi\times \mathbb{R}^{d\times d\times c}\to \mathbb{R}^{d\times d\times c}$ as follows:
\begin{align}
\label{eq:patch}
\mc P(\rho,j_1,j_2)(x)\coloneqq \mathcal R_{d\times d\times c}\left(x\left[j_1:j_1+\rho d,j_2:j_2+\rho d,:\right]\right),
\end{align}
where $(\rho,j_1,j_2)\in \Phi$ and $\mc R_{d\times d\times c}$ resizes the patch to original image size $d\times d\times c$. The patch operator $\mc P$ chooses a quadratic patch of \SI{70}{\percent} to \SI{100}{\percent} the input size and conducts resampling to original size (cf. Figure \ref{fig:model}). 
For resampling, we use bicubic interpolation\footnote{\href{https://www.tensorflow.org/api_docs/python/tf/image/resize}{https://www.tensorflow.org/api\textsubscript{-}docs/python/tf/image/resize}}.

The basic intuition now is: if we take a patch of the input image and propagate it through the generator, than it should be equal to the corresponding patch of the transferred full-size image. 
We choose the 1-norm for comparing the corresponding patches and ensure realistic synthesized patches by adding the patch operator also to the Wasserstein-1 critic.
This yields the following improvements for the critic and the generator risks:

\begin{equation}
\footnotesize
\label{eq:critic2}
\begin{split}
\ell_\text{cri}(\omega,\theta,\mathbf{y},\mathbf{x},p,\vec \phi)\coloneqq \frac{1}{b}\sum_{n=1}^{b}\bigg[&f_\omega(G_\theta(x_n))-f_\omega(y_n)\\
&+f_\omega\left(\mc P_{\phi_n}(G_\theta(x_n))\right)-f_\omega(\mc P_{\phi_n}(y_n))
+p\cdot \left(\ldots\right)^2\bigg],
\end{split}
\end{equation}
\begin{align}
\label{eq:gen2}
\begin{split}
\ell _\text{gen}(\theta,\omega,\mathbf{x},\vec \phi,\lambda)\coloneqq \frac{1}{b}\sum_{n=1}^{b}\bigg[&-f_\omega\left(G_\theta(x_n)\right)-f_\omega\left(\mc P_{\phi_n}\left(G_\theta(x_n)\right)\right)\\
&+\lambda\cdot\underbrace{d^{-2}\norm{G_\theta(\mc P_{\phi_n}(x_n))-\mc P_{\phi_n}(G_\theta(x_n))}_1}_{\ell_\text{patch}(\theta,x_n,\phi_n)}\bigg],
\end{split}
\end{align}
where $\lambda$ controls the influence of the patch loss and the patch extraction settings $\vec \phi=\{\phi_n\}_{n=1}^b,\ \phi_n\in \Phi$ are chosen randomly at each risk calculation.\\

This approach yields some practical advantages: The generator is forced to be consistent over smaller patches, which prevents the network to generate modes with highest similarity to the real data (mode collapse). Furthermore, the generator is prevented from learning arbitrary mappings between input and target domain (e.g., mapping a T1w MRI of an old lady to a T2w MRI of a young boy), because this memorized mapping would then also have to be fulfilled for all smaller patches, i.e., the transfer would also have to be memorized on arbitrary scales. The patch extractor can also be viewed as a magnification and cropping operation. This yields a higher penalty when comparing fine structures that may not have much effect on the loss function when compared on full image scale. Finally, patch extraction causes implicit data augmentation and can help to avoid critic overfitting where the critic is tempted to memorize training samples.

\subsection{Uncertainty by Loss Attenuation}
We consider now Equation \eqref{eq:gen2} from a probabilistic point of view. For $x\in \mc X$, let $a=\mc P_\phi(G_\theta(x)))$ and $b=G_\theta(\mc P_\phi(x))$ for a patch configuration $\phi \in \Phi$. If we force the patch invariance via the 1-norm, the underlying assumption is that every pixel of the residual $\epsilon\coloneqq a-b$ should follow a zero-mean and fixed-scale Laplace distribution \cite{upadhyay2021robustness}. Consider residual pixel $\epsilon_j\sim\text{Laplace}(0,\sigma)(\epsilon_j)=\frac{1}{2\sigma}\exp\left(-| \epsilon_j|/\sigma\right)$ where $\sigma$ represents the scale parameter of the distribution. Maximum likelihood (ML) optimization on the full image (note $a$ and $b$ are functions of $\theta$) yields
\begin{align}
\max_\theta\prod_{j=1}^{d^2}\frac{1}{2\sigma} \exp\left(-\vert a_j-b_j \vert/\sigma\right).
\end{align}
Applying the negative logarithm and dividing by factor $d^2$ results in
\begin{align}
\min_\theta\frac{1}{d^2}\sum_{j=1}^{d^2} \vert a_j-b_j\vert / \sigma + \log(2\sigma),
\end{align}
which is equivalent to minimizing $\ell_\text{patch}(\theta, x, \phi)$ in Equation \eqref{eq:gen2} when considering a fixed scale $\sigma$. The assumption of a fixed scale for the pixel-wise residuals is quite strong and may not hold in the presence of OOD data. The idea now is to consider individual scales for every pixel. Inspired by \cite{upadhyay2021robustness,kendall2017}, we make the scale $\sigma$ a function of input $x$, i.e., we split the generator $G_\theta(x)=[G_\theta^I(x),G_\theta^\sigma(x)]$ in the output branch and return two images, the transferred image $G_\theta^I(x)$ and the corresponding pixel-wise scale map $G_\theta^\sigma(x)$ for the residuals. This results in
\begin{align}
\label{eq:uc}
\ell_\text{patch}(\theta,x,\phi)=\frac{1}{d^2}\sum_{j=1}^{d^2}\frac{|G_\theta(\mc P_{\phi}(x))_j-\mc P_{\phi}(G_\theta(x))_j|}{ G_\theta^\sigma(\mc P_{\phi}(x))_j}+\log\left(2\cdot G_\theta^\sigma(\mc P_{\phi}(x))_j\right).
\end{align}
This can be seen as a loss attenuation as we get high values in $G_\theta^\sigma(x)$ for image regions with high absolute residuals. At the same time, the logarithmic term discourages the model to predict high uncertainty for all pixels. The proposed generator loss for patch invariant and uncertainty-aware image transfer is obtained by inserting $\ell_\text{patch}$ \eqref{eq:uc} into $\ell _\text{gen}$ \eqref{eq:gen2}.

\subsection{Implementation Details}
\label{sec:arch}
In this work, the generator is a U-net \cite{ronneberger2015} with five downsampling operations and approximately \num{10.7e6} parameters. After the last upsampling operation, the U-Net is split into two branches to generate two responses, the transferred image $G_\theta^I(\cdot)$ and the corresponding uncertainty map $G_\theta^\sigma(\cdot)$, c.f.  \eqref{eq:uc}. A non-negative scale map is enforced by applying the softplus activation function ${softplus}(x)\coloneqq\log(\exp(x)+1)$ to the latter output branch. A decoding network for the Wasserstein critic is built following the DCGAN critic \cite{radford2015} with 5 downsampling steps and approximately \num{4.7e6} parameters. Detailed information on critic and generator implementation can be found in the github repository. All models are trained using the Adam optimizer \cite{kingma2014} with $\beta_1=0,\ \beta_2=0.9$ and minibatch size 8. The learning rate is set to \num{5e-5} for the generator and \num{2e-5} for the critic network. No learning rate scheduler or further data augmentation techniques are applied. The total amount of generator updates is 15k and we iterate between 15 critic updates and 1 generator update. Gradient penalty parameter $p$ equals 10, the influence $\lambda$ of the patch constraint is chosen for each data set individually by a grid search.

\section{Experiments}
\subsection{Datasets}
We consider three different tasks in medical image-to-image translation. 

\textbf{Modality Propagation:}
The IXI \cite{ixi} database consists of registered T1w and T2w scans of 577 patients. We want to demonstrate the plausibility of our model for unpaired modality propagation and thus build a model for T1w to T2w transfer. To do so, we remove \SI{10}{\percent} of the patients for evaluation. The remaining patients are randomly split into input and target data, where no patient contributes to both domains at the same time. This is done to simulate a scenario where no paired slices are available throughout the entire training process. We only use the core \SI{60}{\percent} of all axial slices, which yields approximately 20k train slices for input, 20k train slices for target, and 4k pairs for evaluation. The spatial dimension $d$ equals $256$.
	
\textbf{Accelerated MRI enhancement:}
The FastMRI \cite{fastmri} database consists of more than 1500 multi-coil diagnostic knee MRI scans and corresponding emulated single-coil data. Our experiments are based on a subset of nearly 800 coronal proton-density weighted scans without fat-suppression from the official train and validation single-coil releases. We remove \SI{10}{\percent} of the patients for evaluation and split the remaining patients into input and target data. While the target domain consists of slices from fully-sampled MRI scans, we consider 4x acceleration (only \SI{25}{\percent} of k-space measurements) for the slices by using the subsampling scheme discussed in \cite{fastmri,hyun2018}.  This yields 7.1k train slices for input and 6.9k for target. We are aware that enhancement of accelerated MRI can also be considered as a supervised task since generation of paired instances is feasible. However, this experiment should demonstrate possible applicability of the proposed framework to inverse problems in general.

\textbf{MRI-to-CT synthesis}:
The CQ500 \cite{cq500} consists of CT scans of nearly 500 patients, where we use a randomly selected subset of 80 patients for the target domain. Furthermore, we make use of T1w MRI scans of 144 randomly selected patients taken from IXI \cite{ixi} as input data. 
For each dataset we use the core \SI{60}{\percent} of all axial slices, which yields 11.4k train slices for input and 10.4k train slices for target. All slices are subsampled to spatial size $d=256$. Note that in this experiment input and target data is coming from completely separated datasets (inconsistent head orientations, different brain areas, resolution, etc.). 
An additional challenge are artifacts outside the skull caused by the CT table and the measurement equipment in CQ500. We step away from any preprocessing here and investigate how the method reacts to this kind of artifacts.
Since no ground truth data is available for the two separated databases only qualitative evaluation is conducted.

\subsection{Compared Methods \& Scenarios}
We compare our approach to a variety of state-of-the-art methods for unsupervised image transfer that have already been introduced in Section \ref{sec:sota} and Section \ref{sec:uc}: cycle-consistent GAN (cycleGAN) \cite{zhu2017}, uncertainty-aware generalized adaptive cycleGAN (UGAC) \cite{upadhyay2021robustness}, geometry-consistent GAN (gcGAN) \cite{fu2019} with horizontal flip and contrastive unpaired translation (CUT) \cite{park2020}. We test two versions of our approach: the first version utilizes only patch invariance (PI, cf. \eqref{eq:gen2}) and the second version utilizes uncertainty-aware patch invariance (UAPI, cf. \eqref{eq:uc}). For cycleGAN, UGAC and gcGAN we make use of the same generator and critic architecture and training configurations as described in Section \ref{sec:arch} to guarantee a fair comparison. For CUT, we use the publicly available github repository\footnote{\url{https://github.com/taesungp/contrastive-unpaired-translation}}. All slices are normalized to range $[0,255]$ and handled as grayscale images. During optimization, the images are scaled to $[-1,1]$ to speed up training while evaluation metrics, the structural similarity index (SSIM) \cite{ssim} and the peak-signal-to-noise-ratio (PSNR), are calculated on original image scale.

For each of the three applications, we want to test not only performance on unseen assessment data, but also robustness to different types of perturbations. All approaches are trained on unaffected images (without additional noise) and evaluated in the following scenarios: GN0 (original test images); GN5 (adding Gaussian noise, deviation \SI{5}{\percent} of image range); GN10 (adding Gaussian noise, deviation \SI{10}{\percent}); GN20 (adding Gaussian noise, deviation \SI{20}{\percent}); IP2 (impulse perturbation, \SI{2}{\percent} of pixels replaced by random values); IP5 (\SI{5}{\percent} of pixels replaced); IP10 (\SI{10}{\percent} of pixels replaced).

\subsection{Quantitative Evaluation}
\begin{figure}[htb!!]
	\scriptsize
	\begin{floatrow}
		\ttabbox[0.98\columnwidth]{\caption{Quantitative evaluation of our approach and four compared methods on two datasets under seven different noise scenarios. The reported metrics are the structural similarity index and the peak-signal-to-noise-ratio (SSIM/PSNR, higher is better).}}{
			\label{tab:quantitative}
			\begin{tabularx}{.994\columnwidth}{l|c|c|c| c|c|c|c|}
				\toprule
				\textbf{Methods} & \textbf{GN0} & \textbf{GN5} & \textbf{GN10} & \textbf{GN20} & \textbf{IP2} & \textbf{IP5} & \textbf{IP10}
				\\ \midrule
				
				&\multicolumn{7}{c|}{IXI test data}\\ \midrule
				cycleGAN&$78.99 / 20.90$&$77.04 / 20.71$&$72.53 / 20.14$&$63.87 / 19.19$&$69.81 / 20.39$&$64.51 / 19.99$&$60.08 / 19.50$\\ \midrule
				UGAC&$79.31 / 21.21$&$77.24 / 21.05$&$72.53 / 20.43$&$\textbf{64.03} / 19.37$&$70.15 / 20.68$&$64.88 / 20.22$&$61.01 / 19.78$\\ \midrule
				gcGAN&$81.00 / 21.52$&$76.51 / 20.66$&$69.96 / 19.50$&$57.75 / 17.90$&$73.79 / 20.30$&$69.24 / 19.55$&$63.91 / 18.78$\\ \midrule
				CUT&$78.67 / 21.15$&$76.10 / 21.22$&$71.43 / 20.58$&$61.27 / 19.67$&$74.11 / 21.19$&$71.78 / 20.94$&$68.97 / 20.65$\\ \midrule
				PI(ours)&$\textbf{82.53} / 22.18$&$69.82 / 21.49$&$51.90 / 20.64$&$36.96 / 19.26$&$54.85 / 20.92$&$44.10 / 20.24$&$38.73 / 19.70$\\ \midrule
				UAPI(ours)&$79.99 / \textbf{22.62}$&$\textbf{78.76} / \textbf{22.20}$&$\textbf{74.49} / \textbf{21.70}$&$62.24 / \textbf{20.34}$&$\textbf{77.44} / \textbf{21.93}$&$\textbf{74.29} / \textbf{21.56}$&$\textbf{69.35} / \textbf{21.14}$\\ \midrule \midrule
				
				&\multicolumn{7}{c|}{FastMRI test data}\\ \midrule
				cycleGAN&$81.60 / 21.48$&$70.71 / 19.45$&$59.59 / 17.89$&$40.87 / 15.80$&$62.95 / 18.35$&$57.50 / 17.74$&$50.80 / 17.02$\\ \midrule
				UGAC&$85.29 / 22.62$&$72.17 / 19.91$&$59.64 / 17.99$&$43.21 / 16.24$&$72.01 / 20.03$&$63.97 / 18.73$&$56.36 / 17.71$\\ \midrule
				gcGAN&$80.73 / 20.76$&$68.67 / 18.13$&$59.23 / 16.60$&$48.31 / 15.26$&$71.51 / 18.61$&$65.22 / 17.54$&$58.66 / 16.61$\\ \midrule
				CUT&${89.36} / 23.37$&$\textbf{85.67} / \textbf{22.71}$&$\textbf{83.32} /\textbf{ 22.15}$&$\textbf{76.63} /\textbf{ 20.58}$&$\textbf{86.61} / \textbf{23.02}$&$\textbf{84.45} /\textbf{ 22.45}$&$\textbf{82.59 }/ \textbf{21.86}$\\ \midrule
				PI(ours)&$85.85 / 23.37$&$76.22 / 20.85$&$66.44 / 18.94$&$53.46 / 17.19$&$75.77 / 20.80$&$67.61 / 19.28$&$59.44 / 18.04$\\ \midrule
				UAPI(ours)&$\textbf{90.37} /\textbf{ 24.74}$&$81.08 / 21.37$&$69.97 / 18.41$&$55.72 / 16.07$&$79.18 / 19.81$&$71.69 / 17.97$&$65.44 / 16.98$\\ \bottomrule
			\end{tabularx}
		}
	\end{floatrow}
\end{figure}
\begin{figure}[htb!]
	\centering
	\includegraphics[width=.99\columnwidth]{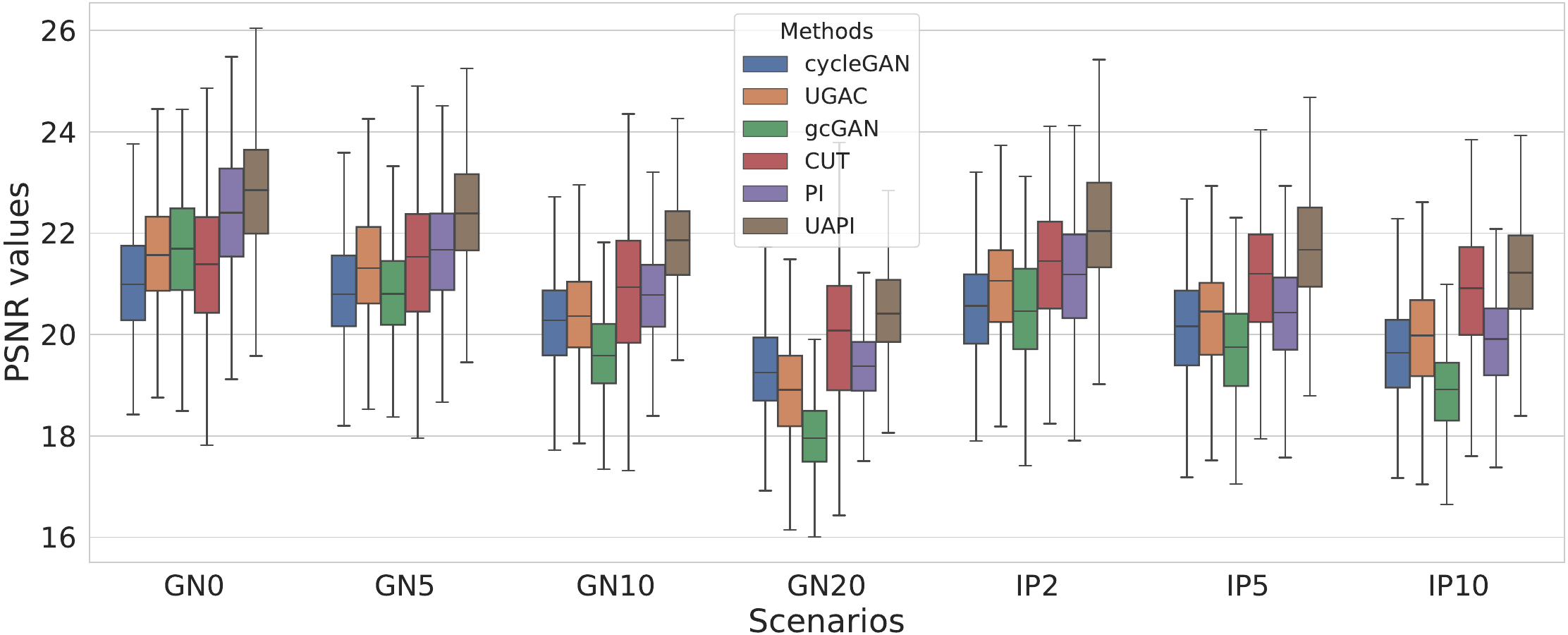}
	\caption{Visual analysis of the PSNR values on IXI under seven different test scenarios, obtained by our two approaches (PI, UAPI) and 4 compared methods (cycleGAN, UGAC, gcGAN, CUT).}
	\label{fig:boxixi}
\end{figure}

The quantitative metrics in Table \ref{tab:quantitative} obtained on unseen test data indicate superior performance of our approach for the modality propagation task on IXI. Considering evaluation on clean test data, usage of patch invariance gives an increase in SSIM and PSNR metrics compared to the bi-directional (cycleGAN, UGAC) and uni-directional (gcGAN, CUT) benchmarks.
 As compared to the benchmark methods, consideration of uncertainty-aware patch invariance  even improves results, also for the scenarios with perturbed test data (GN5 to IP10).
  Robustness of UAPI is also established visually in Figure \ref{fig:boxixi}. 
 Especially for scenarios with high perturbations (GN10, GN20, IP5, IP10), we observe a performance advandtage when using the uncertainty-aware method UAPI.
 Interestingly, usage of PI without uncertainty awareness on perturbation scenarios GN5 to IP10 yields a significant decrease for the SSIM but not for the PSNR metric. This rather unexpected observation needs further investigation in future research.

\begin{figure}[htb!]
	\centering
	\includegraphics[width=.99\columnwidth]{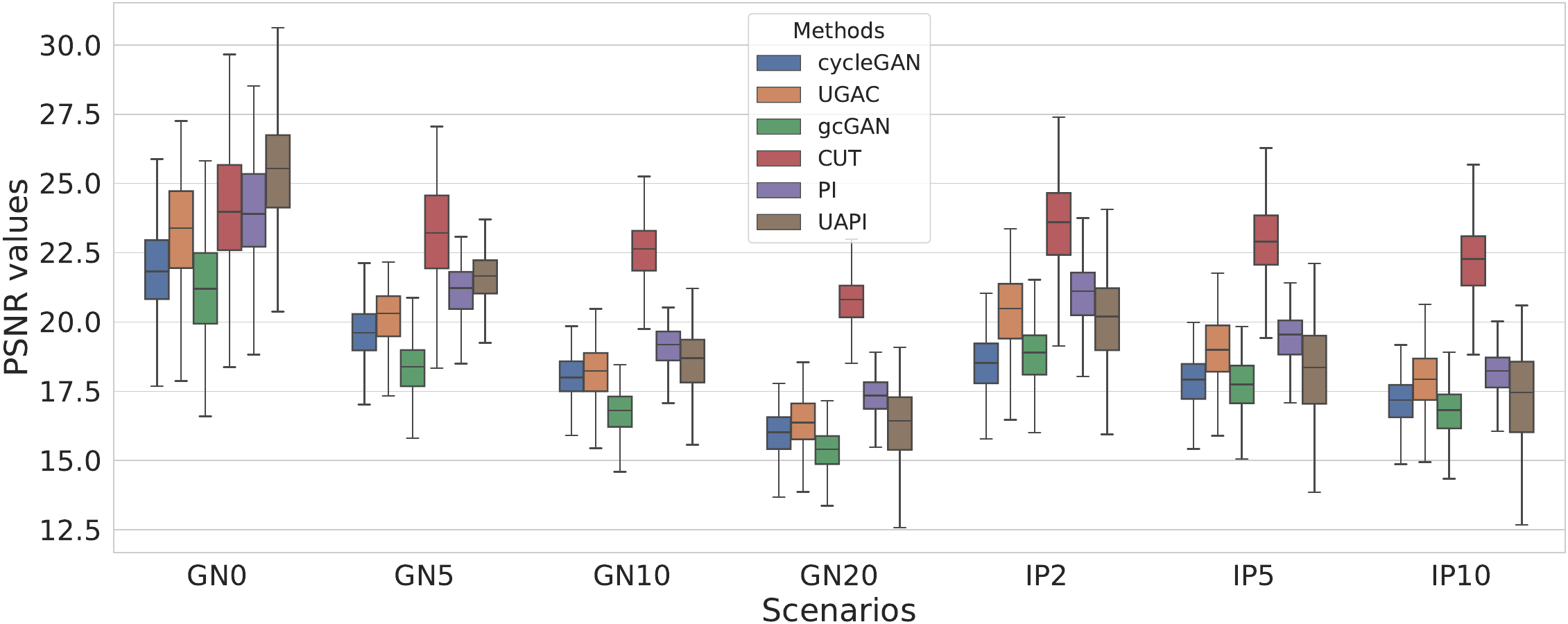}
	\caption{Visual analysis of the PSNR values on FastMRI under seven different test scenarios, obtained by our two approaches (PI, UAPI) and 4 compared methods (cycleGAN, UGAC, gcGAN, CUT).}
	\label{fig:boxkne}
\end{figure}

In Table \ref{tab:quantitative} we observe for the accelerated MRI enhancement task on FastMRI that our method UAPI significantly outperforms other benchmark on unaffected test data but gives modest accuracy when additional noise and perturbed pixels are added. Figure \ref{fig:boxkne} shows the superior performance of CUT in terms of the PSNR metric for noisy data. This is quite interesting since that has not been the case for the previous modality propagation application. The task of accelerated MRI enhancement strongly differs from the other two applications. While the goal of modality propagation and MRI-to-CT synthesis is to come up with a completely new image, the aim of accelerated MRI enhancement is to improve quality of a already existing image. 
 In fact, the methods cycleGAN, UGAC, gcGAN, PI and UAPI depend on a rather simple U-Net \cite{ronneberger2015} implementation and a standard DCGAN critic \cite{radford2015} with the aim to demonstrate plausibility of different transfer approaches on easy-to-implement frameworks. The CUT method is a benchmark where the publicly available source code had to be used, consisting of a ResNet-based generator \cite{zhu2017} and built-in data augmentation techniques that may better compensate for noisy input data. Nevertheless, our methods PI and UAPI  seemingly achieve better results compared to the U-Net based benchmarks. We will take up investigation of robustness of our methods in combination with different network architectures as a future goal.
 
\subsection{Qualitative Evaluation}
\begin{figure}[htb!]
	\centering
	\includegraphics[width=0.12\columnwidth]{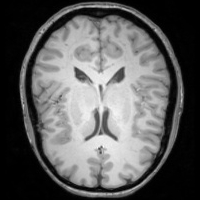}\quad
	\includegraphics[width=0.12\columnwidth]{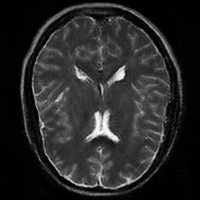}\hfill
	\includegraphics[width=0.12\columnwidth]{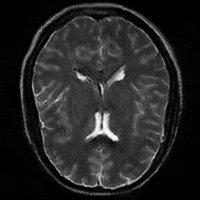}\hfill
	\includegraphics[width=0.12\columnwidth]{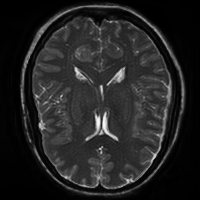}\hfill
	\includegraphics[width=0.12\columnwidth]{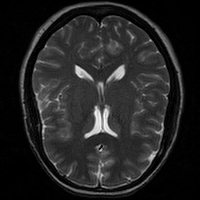}\hfill
	\includegraphics[width=0.12\columnwidth]{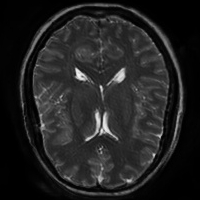}\hfill
	\includegraphics[width=0.12\columnwidth]{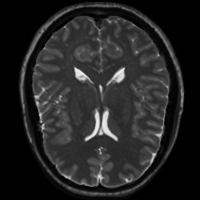}\hfill
	\includegraphics[width=0.12\columnwidth]{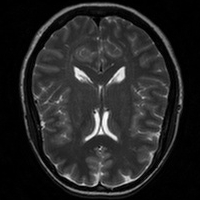}
	
	\includegraphics[width=0.12\columnwidth]{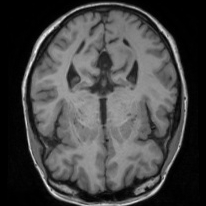}\quad
	\includegraphics[width=0.12\columnwidth]{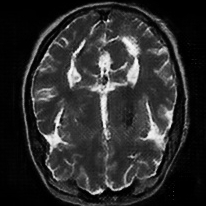}\hfill
	\includegraphics[width=0.12\columnwidth]{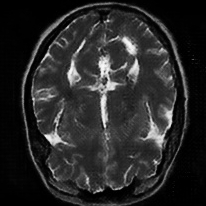}\hfill
	\includegraphics[width=0.12\columnwidth]{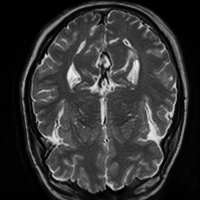}\hfill
	\includegraphics[width=0.12\columnwidth]{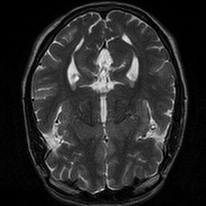}\hfill
	\includegraphics[width=0.12\columnwidth]{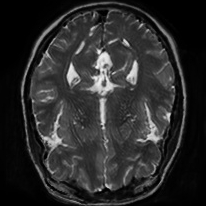}\hfill
	\includegraphics[width=0.12\columnwidth]{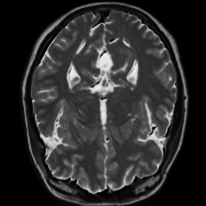}\hfill
	\includegraphics[width=0.12\columnwidth]{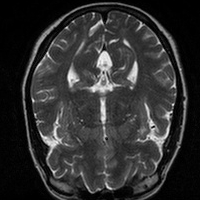}\\[.4em]
	
	\includegraphics[width=0.12\columnwidth]{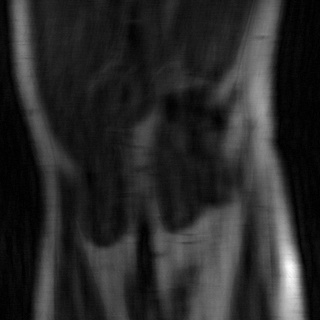}\quad
	\includegraphics[width=0.12\columnwidth]{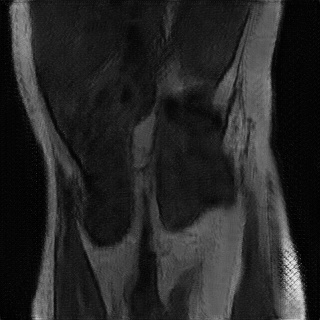}\hfill
	\includegraphics[width=0.12\columnwidth]{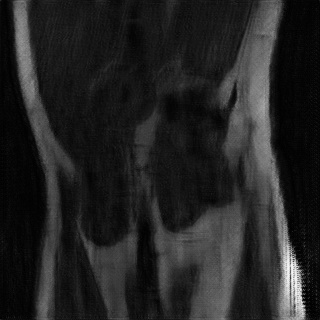}\hfill
	\includegraphics[width=0.12\columnwidth]{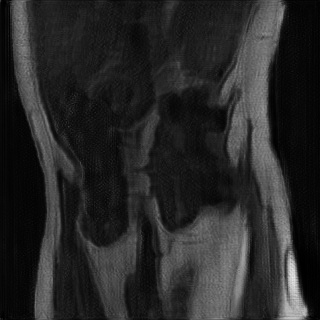}\hfill
	\includegraphics[width=0.12\columnwidth]{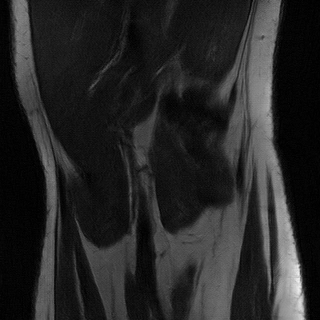}\hfill
	\includegraphics[width=0.12\columnwidth]{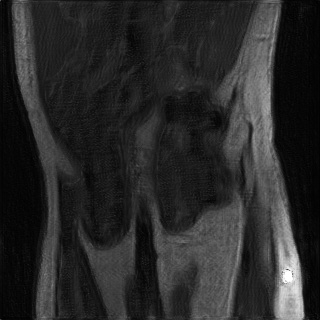}\hfill
	\includegraphics[width=0.12\columnwidth]{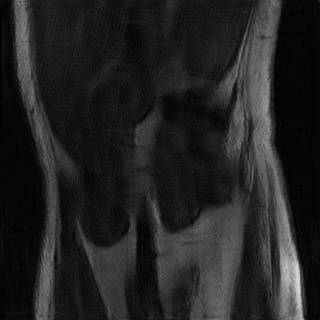}\hfill
	\includegraphics[width=0.12\columnwidth]{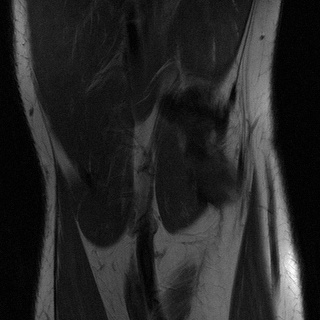}
	
	\includegraphics[width=0.12\columnwidth]{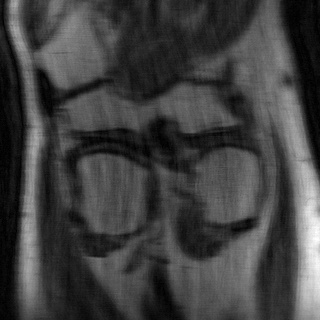}\quad
	\includegraphics[width=0.12\columnwidth]{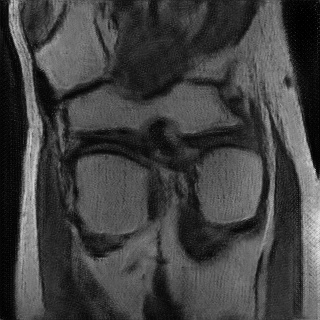}\hfill
	\includegraphics[width=0.12\columnwidth]{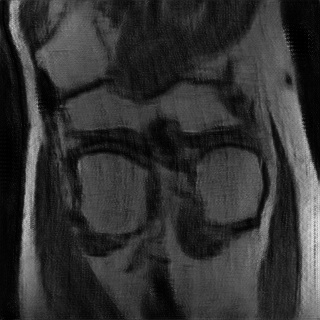}\hfill
	\includegraphics[width=0.12\columnwidth]{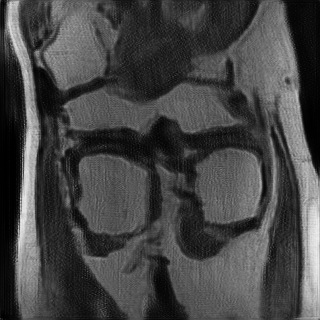}\hfill
	\includegraphics[width=0.12\columnwidth]{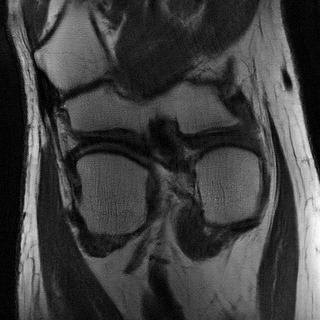}\hfill
	\includegraphics[width=0.12\columnwidth]{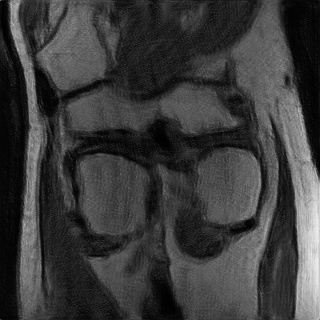}\hfill
	\includegraphics[width=0.12\columnwidth]{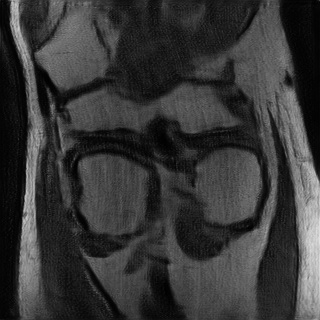}\hfill
	\includegraphics[width=0.12\columnwidth]{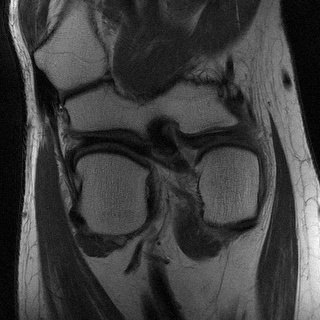}

	\caption{Evaluation samples of our approach and four compared methods on IXI and FastMRI for scenario GN0 (test data without perturbations). From left to right: input, images transferred by cycleGAN, UGAC, gcGAN, CUT, PI, UAPI and ground truth.}
	\label{fig:qualitative}
\end{figure}
 In Figure \ref{fig:qualitative} we analyze the prediction quality of our and compared approaches in a qualitative way. Considering modality propagation in MRI, we see that usage of uncertainty-aware patch invariance (UAPI) gives a better detailed weighting of the cerebrospinal fluid in the middle of the brain. In general, employing patch invariance yields better preservation of fine structures.
This observation also applies to accelerated MRI enhancement. In particular, CUT and UAPI provide comparatively sharper knee images with more high-frequency details than the other methods.
 
\begin{figure}[htb!]
	\centering
	\includegraphics[width=\columnwidth]{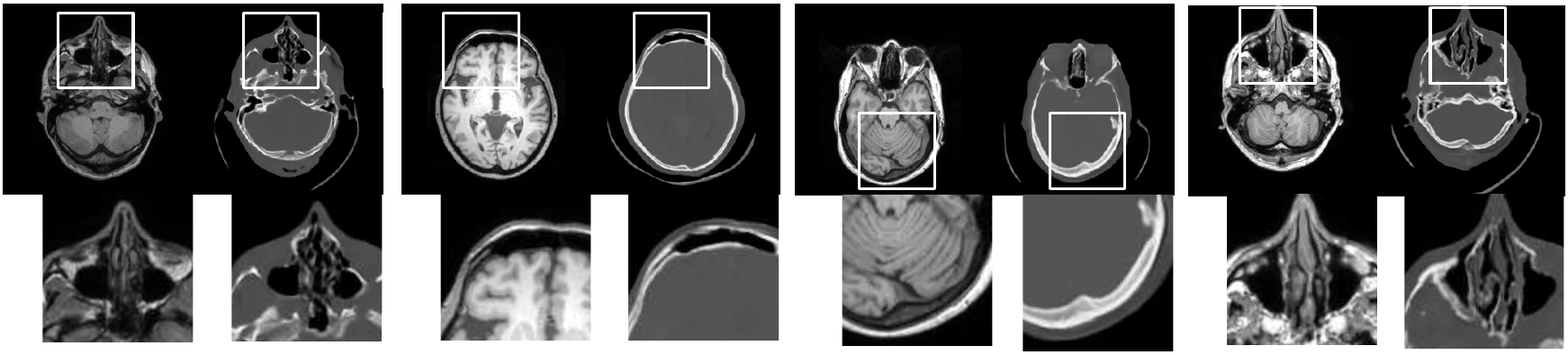}
	\caption{Evaluation samples of the UAPI method on unseen brain MRI slices. For every data pair, the input slice and the corresponding UAPI prediction are visualized on the left and the right side, respectively. The first row contains the images on original scale, the second row selected patches to visualize the prediction quality for detailed structures.}
		\label{fig:hea}
\end{figure}

Qualitative evaluation plays an important role for the third investigated application, namely MRI-to-CT synthesis, where quantitative comparison is not possible due to lack of ground truth data. Satisfying results were obtained with the UAPI method, which are visualized in Figure \ref{fig:hea}. Cavities and brain shapes are well preserved by our method although we used two completely independent and unaligned head datasets for this experiment. UAPI synthesizes brain table artifacts that are also visible in CQ500. A proper evaluation on cleaned CT data is necessary and thus will be considered as a future working step.

\subsection{Uncertainty Scores}
\begin{figure}[htb!]
	\centering
	\begin{floatrow}
		\ffigbox[.52\columnwidth]{
			
			\includegraphics[width=1.02\columnwidth]{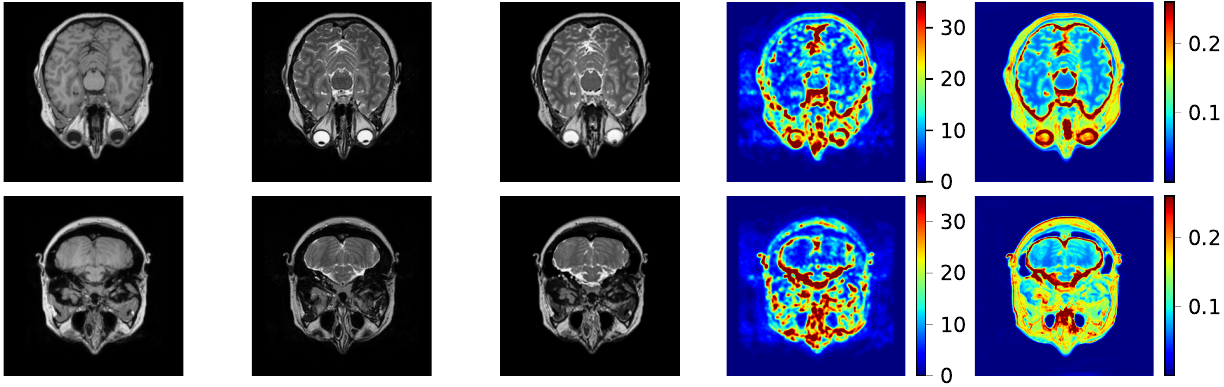}\\[1.9em]
			\includegraphics[width=1.02\columnwidth]{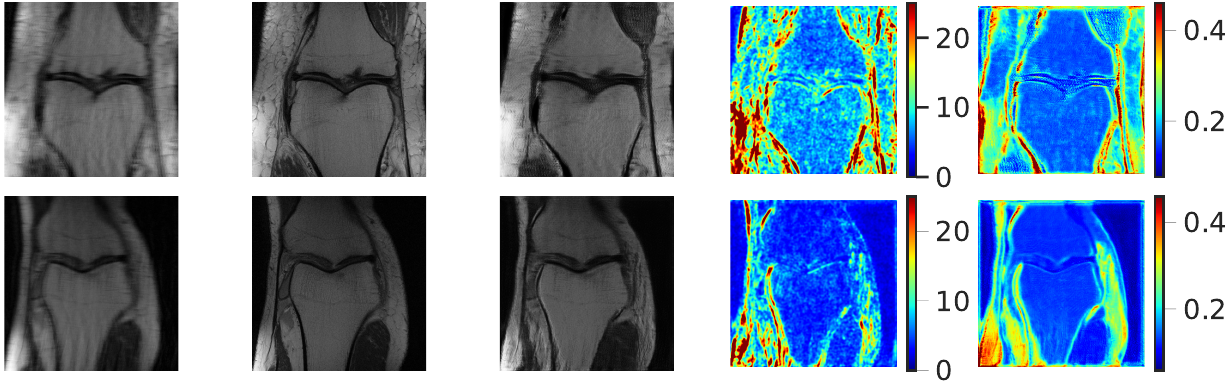}
		}{
			\caption{Position-based relation between abs. residuals and predicted scale maps on IXI (top) and FastMRI (bottom). Left to right: input, ground truth, prediction by UAPI, abs. residuals and predicted scale map.}\label{fig:ixiuc1}
		}\hfill
		\ffigbox[.44\columnwidth]{
			\includegraphics[width=.49\columnwidth]{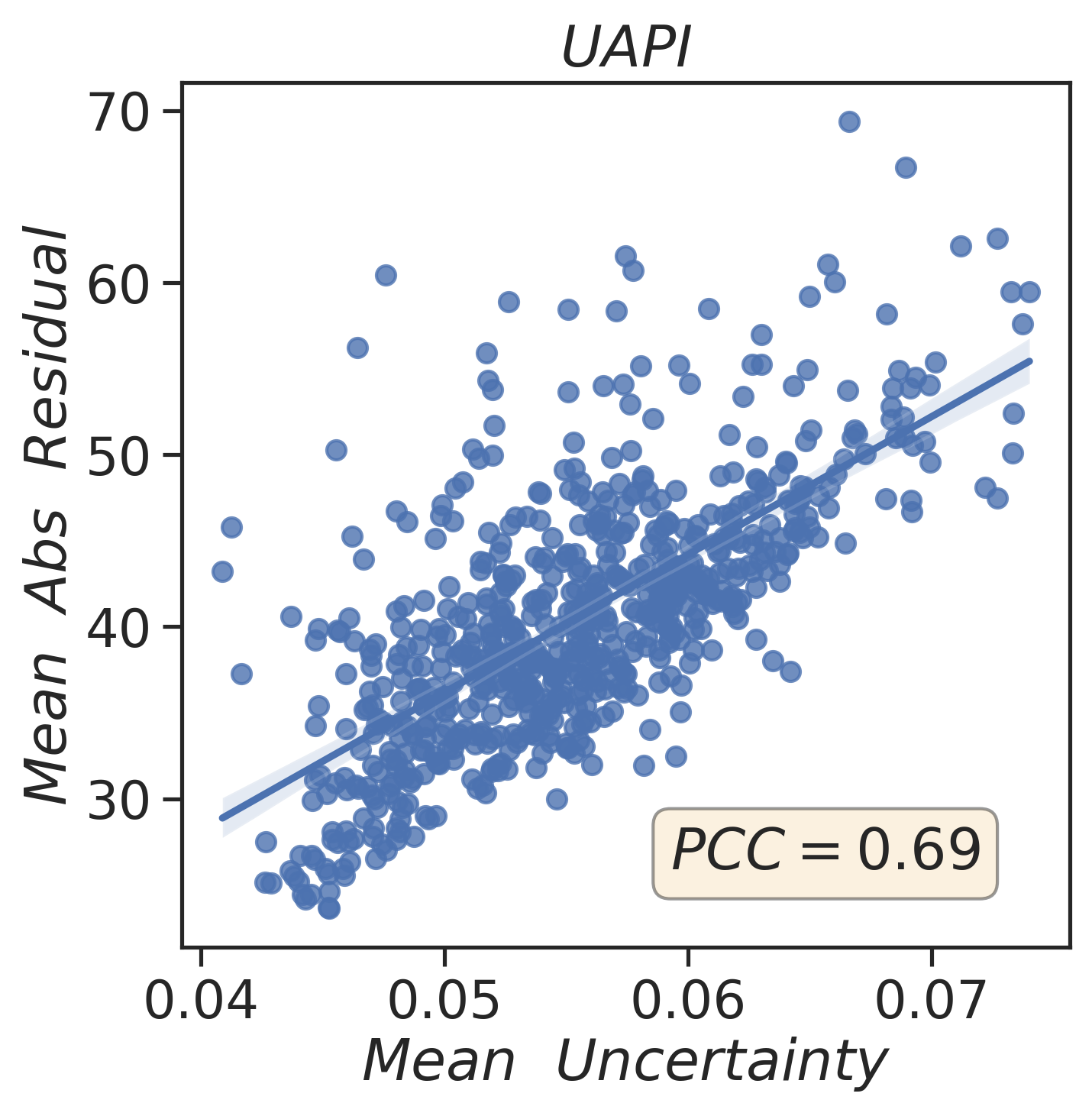}\hfill
			\includegraphics[width=.49\columnwidth]{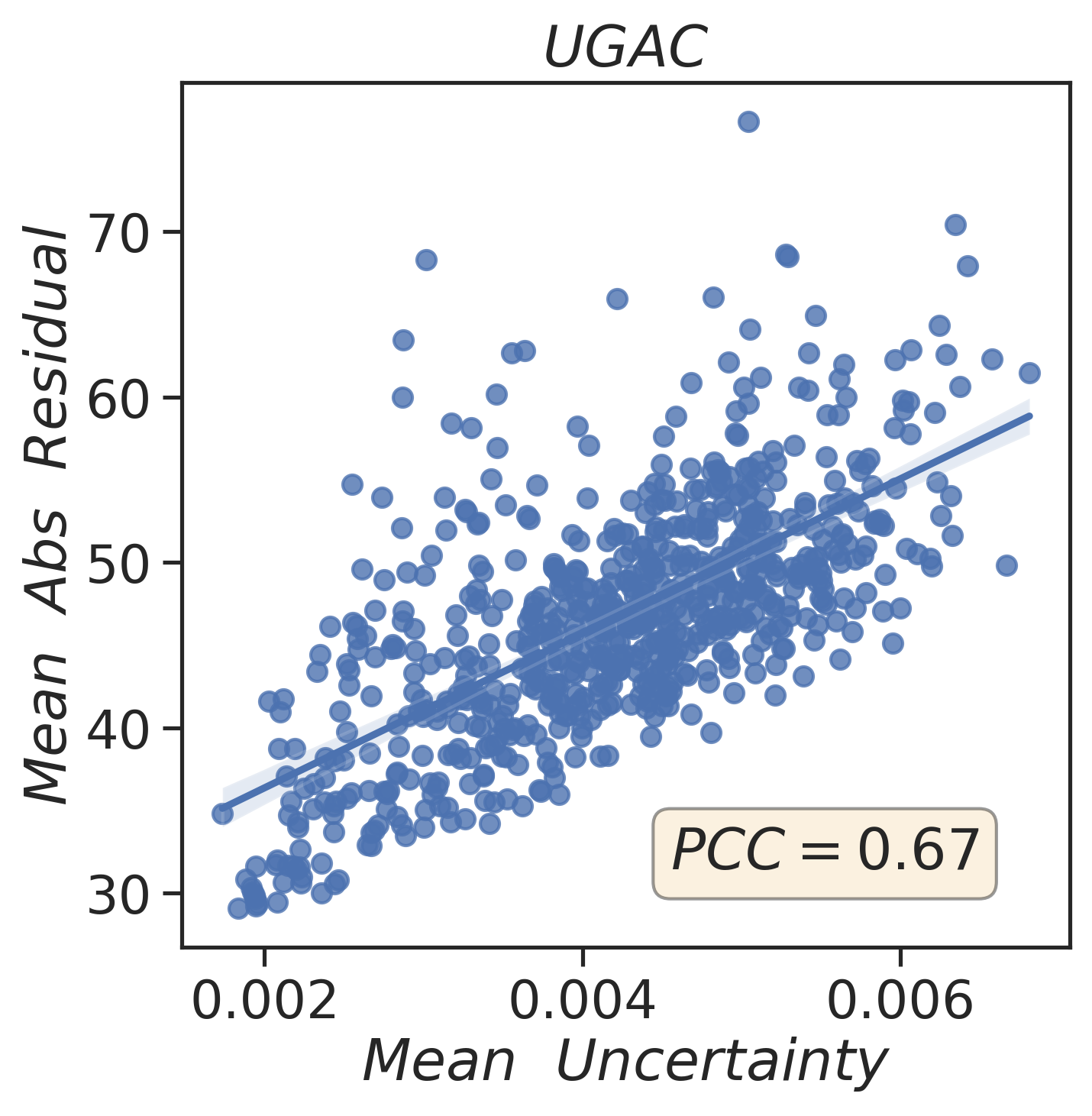}\\[.2em]
				\includegraphics[width=.49\columnwidth]{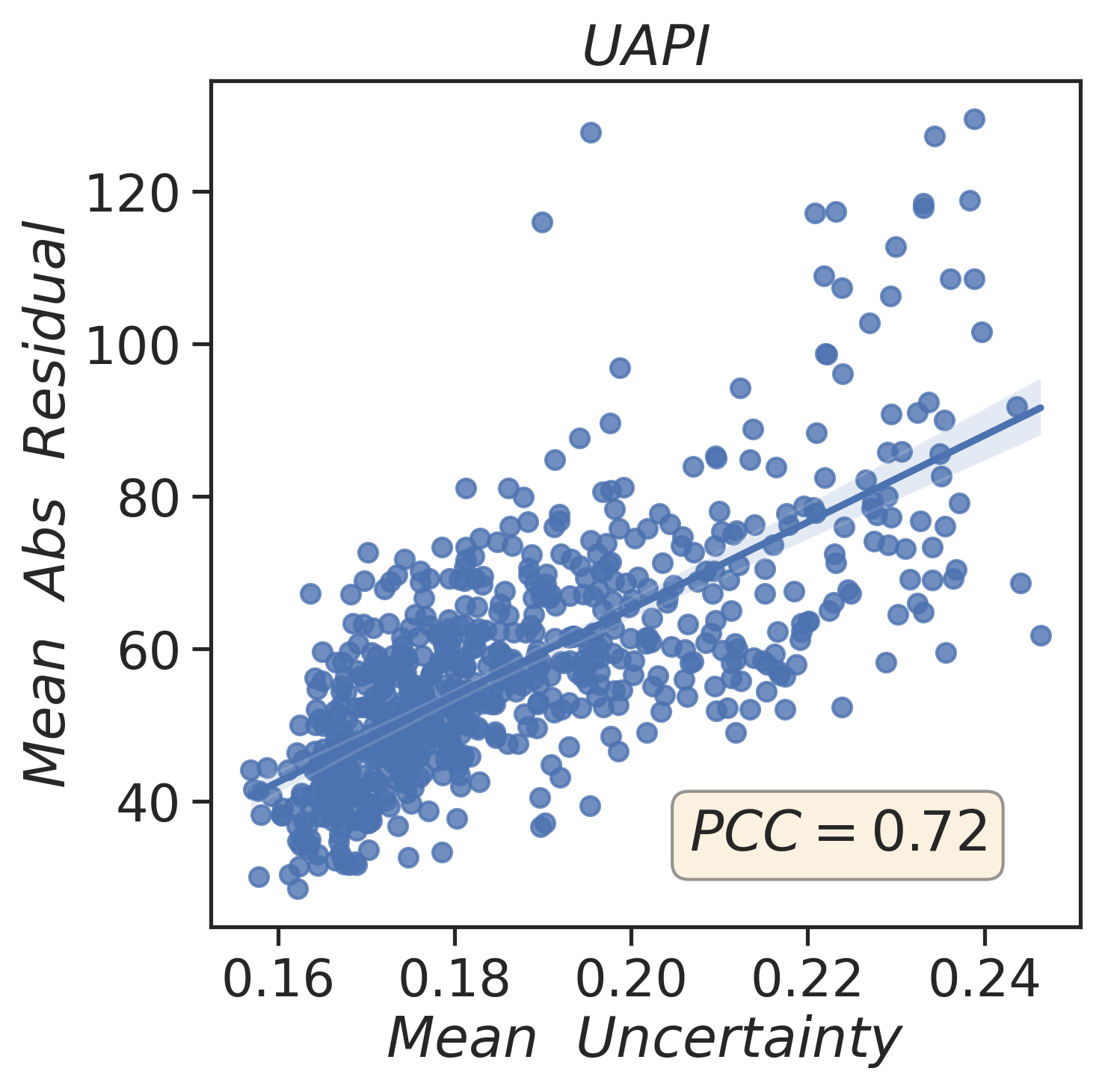}\hfill
			\includegraphics[width=.49\columnwidth]{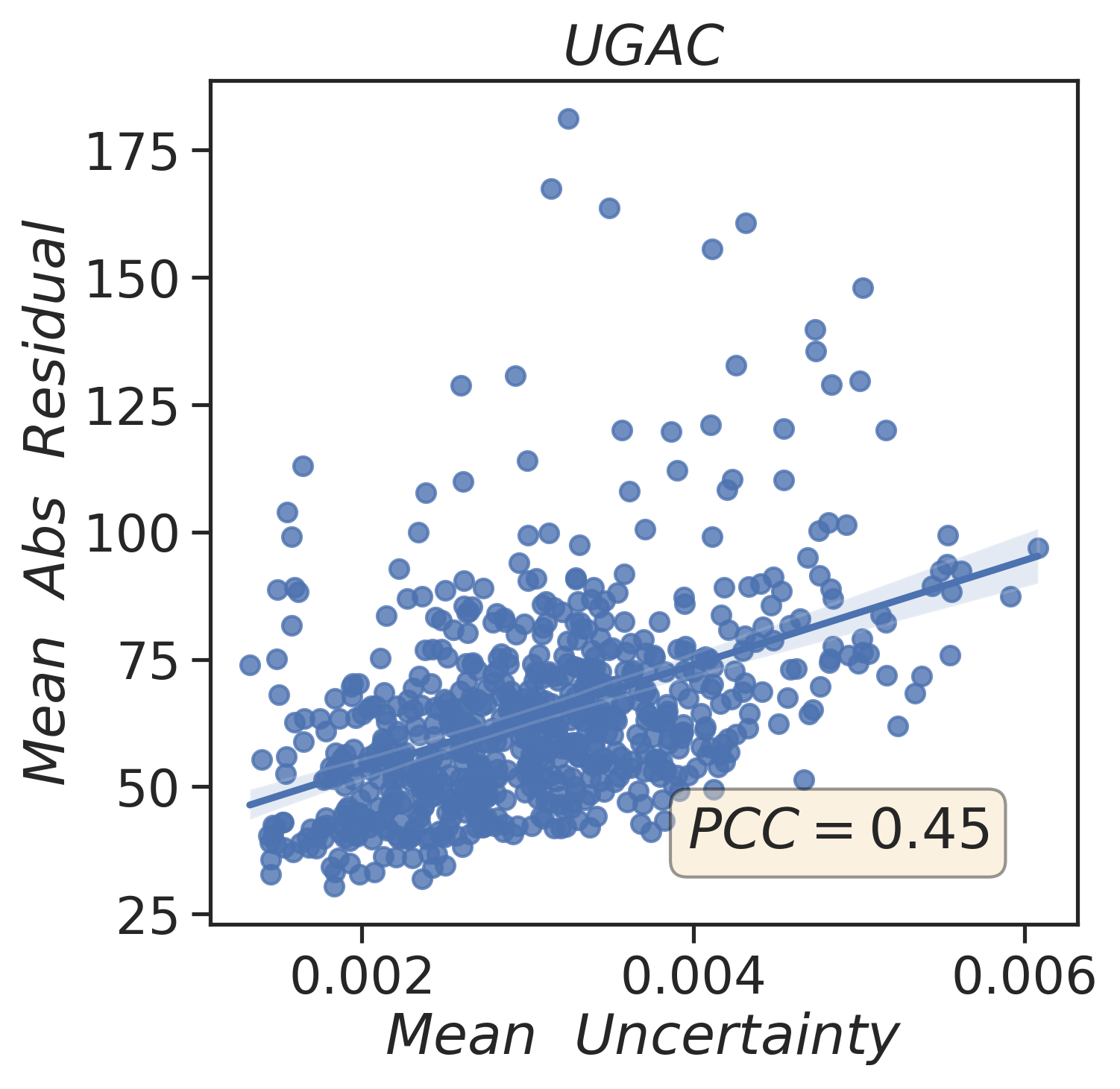}
		}{\caption{Scatter plot between abs. residual and scale map values on IXI (top) and FastMRI (bottom). The predictions are generated by UAPI (left) and UGAC (right).}\label{fig:scatterixi}}
	\end{floatrow}
\end{figure}

Additional to improved accuracy we demonstrate the efficacy of estimating the scale maps with the proposed method. The input-dependent non-negative scale maps are derived from the second output branch $G_\theta^\sigma$, see \eqref{eq:uc}. Indeed, the predicted scale maps are able to model uncertainty inherent from data. This can be observed in Figure \ref{fig:ixiuc1}, where in addition to the transferred images also the predicted scale maps and the absolute residuals between predicted and ground truth images are displayed. Obviously, uncertainty is relatively greater in regions with higher residual values. From the scale maps it can be deduced for which positions the generator is comparatively uncertain in its prediction, such as the cerebral cortex and eye sockets in head MRI or the lateral knee ligaments in knee MRI.

The correspondence between residual and scale maps suggests that the latter can be used as an approximation to a prediction's residuals that are not available due to the lack of ground truth data in unsupervised learning. In order to quantitatively study this relationship we visualize mean absolute residual score and mean uncertainty maps for 512 randomly selected unseen test images in a scatter plot (see Figure \ref{fig:scatterixi}). Moreover, we compare our uni-directional method UAPI also to the relations observed by UGAC that models uncertainty with the help of a bi-directional cycleGAN \cite{upadhyay2021robustness}. For modality propagation as well as accelerated MRI enhancement we visually observe an approximate positive linear correlation between mean absolute residual scores and mean uncertainty scores. We calculate the Pearson correlation coefficient (PCC) to obtain a quality estimate for the linear correlation and compare between UAPI and UGAC. Our method returns a slightly higher PCC on IXI (UAPI: 0.69, UGAC: 0.67). The discrepancy between both methods even increases on FastMRI (UAPI: 0.72, UGAC: 0.45). This further encourages the idea that scale maps derived from our approach can be used to indicate the overall quality of a transferred image.

\section{Conclusions}
In this paper we proposed a WGAN-based approach using patch invariance to employ joint image transfer and uncertainty quantification in an fully unsupervised manner. We demonstrate superior performance of our uni-directional method for modality propagation and accelerated MRI enhancement compared to four state-of-the-art benchmarks in unpaired image translation. 
Moreover, the method reaches qualitatively satisfying results for MRI-to-CT synthesis using completely unaligned databases during training. 
The predicted uncertainty can be representative of the residual maps and thus indicate the quality of a transferred image in the absence of ground truth data. 
Further investigation of the network architecture and improvement in robustness represents an important goal for future research.
Future work will also include the application of the uncertainty-aware and patch invariant network to other unpaired image-to-image applications outside the medical sector.

\clearpage
%
%
\bibliographystyle{splncs03_unsrt}
\bibliography{egbib}
\end{document}